\documentclass[preprint,12pt, authoryear]{elsarticle}
\usepackage{amssymb}
\usepackage{amsmath}
\usepackage{makecell}
\usepackage{url}
\usepackage{dsfont}
\usepackage[hidelinks]{hyperref}  
\usepackage{makecell}
\usepackage{xcolor}
\usepackage{float}
\usepackage{breakurl}
\journal{Neural Networks}

\makeatletter
\def\ps@pprintTitle{
 \let\@oddhead\@empty
 \let\@evenhead\@empty
 \let\@oddfoot\@empty
 \let\@evenfoot\@empty
}
\makeatother

\begin{document}
\sloppy

\begin{frontmatter}

\title{General Transform: A Unified Framework for Adaptive Transform to Enhance Representations}

\author[label1,label2]{Gekko Budiutama}
\corref{cor1}
\ead{bgekko@quemix.com}
\cortext[cor1]{Corresponding author}
\author[label3]{Shunsuke Daimon}
\author[label1,label2]{Hirofumi Nishi}
\author[label1,label2,label3]{Yu-ichiro Matsushita}

\affiliation[label1]{organization={Quemix Inc.}, 
            city={Tokyo}, 
            postcode={1030027}, 
            country={Japan}}

\affiliation[label2]{organization={The University of Tokyo}, 
            orgdiv={Department of Physics}, 
            city={Tokyo}, 
            postcode={1130033}, 
            country={Japan}}

\affiliation[label3]{organization={National Institutes for Quantum Science and Technology}, 
            orgdiv={Quantum Materials and Applications Research Center}, 
            city={Tokyo}, 
            postcode={1528550}, 
            country={Japan}}

\begin{abstract}
Discrete transforms, such as the discrete Fourier transform, are widely used in machine learning to improve model performance by extracting meaningful features. However, with numerous transforms available, selecting an appropriate one often depends on understanding the dataset’s properties, making the approach less effective when such knowledge is unavailable. In this work, we propose General Transform (GT), an adaptive transform-based representation designed for machine learning applications. Unlike conventional transforms, GT learns data-driven mapping tailored to the dataset and task of interest. Here, we demonstrate that models incorporating GT outperform conventional transform-based approaches across computer vision and natural language processing tasks, highlighting its effectiveness in diverse learning scenarios.
\end{abstract}




\end{frontmatter}


\section{Introduction}\label{sec1}

Deep neural networks have consistently pushed the boundaries of performance on tasks in computer vision, natural language processing, and beyond. A significant trend in improving the efficiency of these systems involves the integration of discrete mathematical transforms, such as the discrete Fourier transform, discrete cosine transform, and others \citep{https://doi.org/10.1049/sil2.12109, Yi2023ASO}.

The use of discrete transforms provides multiple advantages in deep learning, particularly in handling high-dimensional data. These transforms enable efficient signal compression and decorrelation, which are crucial for a wide range of data types. The discrete cosine transform has been widely employed in image compression schemes, such as JPEG, due to its ability to concentrate energy into a few significant coefficients \citep{DCT-defined, DCT-applied}. Similarly, the fast Fourier transform is fundamental in spectral analysis and audio compression, as seen in MP3 encoding and speech processing \citep{fft}. The discrete wavelet transform further extends these benefits by allowing multi-resolution analysis, making it highly effective in image compression formats like JPEG 2000 \citep{wavelet}. Beyond compression, discrete transforms may also contribute to improved generalization in deep learning by filtering high-frequency noise and emphasizing dominant spectral components for computer vision tasks \citep{Xu2020}. Additionally, frequency-domain representations improve computational efficiency, as certain operations—such as convolutions—can be performed more efficiently in the spectral domain. For example, Fourier-based convolutional architectures leverage the fast Fourier transform to replace spatial-domain convolutions with element-wise multiplications, significantly reducing computational complexity \citep{Chitsaz2020AccelerationOC, Highlander2016VeryET}.

Despite their advantages, a fundamental limitation of these approaches is their reliance on predefined operations. The chosen transform applies the same mapping regardless of the input's specific characteristics, thus may fail to optimally capture task-specific or domain-specific features. Moreover, selecting an appropriate transform often requires prior knowledge of the dataset. This restricts generality and reduces the suitability of these methods for multimodal or highly heterogeneous data.

To address these challenges, we introduce General Transform (GT), a parameterized approach enabling flexible adaptation across various transforms. GT dynamically learns the most suitable transform or combination of transforms directly from task and dataset. The additional parameters introduced by GT is directly proportional to the number of included transforms, which remains relatively insignificant compared to the overall network size. Here, we demonstrate the application of GT in replacing conventional transforms for computer vision and natural language processing tasks. We found that the proposed GT enhances performance in large models while adding only three additional parameters. Although our proof-of-concept focuses on image and text classification tasks, GT is a general approach that can be extended to other applications, making it a versatile tool across various domains. Thus, our main contributions include:
\begin{itemize}
    \item General Transform (GT) is introduced as a parameterized approach that dynamically adapts to different transforms, learning the most suitable transforms directly from the task and dataset.
    \item The impact of GT on enhancing the performance of large deep learning models is analyzed for computer vision and natural language tasks, requiring only three additional parameters.
    \item GT is proven to be capable of capturing meaningful differences across different input channels in both image and language data processing.
\end{itemize}

\section{Related Works}

Discrete transforms have become increasingly prominent in machine learning, offering efficient methods to manipulate and analyze data across various domains. From computer vision to natural language processing (NLP) and time series analysis, these mathematical tools enable both improved performance and reduced computational overhead. Below, we provide an in-depth exploration of the most widely used discrete transforms in machine learning, accompanied by their key applications and recent research trends.

\subsection{Discrete Fourier Transform (DFT)}

DFT is defined as: 
\begin{gather}
    X[k] = \sum_{n=0}^{N-1} e^{-\frac{j2\pi kn}{N}}\, x[n],
\end{gather}
where $n$ and $k$ are indices, $x[n]$ represents the original sequence of length $N$, $X[k]$ is the transformed representation of $x[n]$, underpins many frequency-domain methods in machine learning. In computer vision, the introduction of Fourier Neural Operators \citep{guibas2022adaptivefourierneuraloperators} uses DFT to carry out global operations in the Fourier space, improving both model generalization and efficiency. Similarly, DFT-based positional encodings \citep{tancik2020fourier} help capture high-frequency details within vision transformer architectures.

In NLP, FNet \citep{leethorp2022fnetmixingtokensfourier} exemplifies how replacing self-attention in the Transformer architecture \citep{VaswaniAttention} with Fourier-based token mixing can significantly reduce computational complexity while preserving strong performance. Autoregressive models also benefit from frequency-domain mixing \citep{lou2021fnetarmixingtokensautoregressive}, facilitating more efficient sequence modeling.

Time series analysis has leveraged DFT to model periodic behavior. TimeMixer \citep{wang2024timemixer}, for instance, employs Fourier transforms to capture extended temporal dependencies, resulting in more robust forecasting. Additionally, model merging can be streamlined by transforming network parameters via DFT, as shown in \citep{zheng2024freemergingfouriertransformmodel}, while \citep{DFTFineTuning} demonstrated that the Discrete Fourier Transform (DFT) can be leveraged for parameter-efficient fine-tuning of large language models.

\subsection{Discrete Cosine Transform Type-II (DCT)}

DCT, given by:
\begin{gather}
    X[k] = \sum_{n=0}^{N-1} \cos\Bigl(\tfrac{\pi k \,(n+0.5)}{N}\Bigr)\, x[n],
\end{gather}
is widely recognized for decorrelation and compression, making it pivotal in image processing pipelines (e.g., JPEG) \citep{JPEG}. Recent work in computer vision \citep{Xu2020, 7415375, Karaoglu2023, Lee2024, SU2024106139} integrates DCT features or DCT-based attention, enhancing image classification and feature extraction. DCT-based decorrelated attention \citep{pan2024dctbaseddecorrelatedattentionvision} has further demonstrated computational gains over conventional self-attention approaches.

In NLP, DCT-derived spectral token representations \citep{Scribano2023} have shown promise for various tasks, including sentiment analysis and machine translation, by emphasizing important frequency components of textual data.

\subsection{Discrete Wavelet Transform (DWT)}

The Haar DWT in pure element–wise form is:
\begin{gather}
  X[k] \;=\; \sum_{n=0}^{N-1} H(k,n)\, x[n]\,, 
  \quad k = 0,\dots,N-1,
\end{gather}
where \(N=2^J\), and the scalar‐kernel \(H(k,n)\) is
\begin{gather}
  H(k,n)=
  \begin{cases}
    2^{-J/2}, 
    & k = 0,\\[0.5em]
    2^{-\,j/2}, 
    & k = 2^{\,J-j} + m,\;\;2^j m \le n < 2^j m + 2^{\,j-1},\\[0.3em]
    -2^{-\,j/2}, 
    & k = 2^{\,J-j} + m,\;\;2^j m + 2^{\,j-1} \le n < 2^j(m+1),\\[0.3em]
    0, & \text{otherwise},
  \end{cases}
\end{gather}
for levels \(j=1,\dots,J\) and shifts \(m=0,\dots,2^{\,J-j}-1\).  

Haar-based DWTs are used in deep learning pipelines to improve spatial-frequency modeling. For instance, wavelet CNNs \citep{liu2019multi} apply DWT to decompose inputs into multiple resolutions before convolution. In generative modeling, WaveletFlow \citep{WaveletFlow} uses DWT for reversible transformations that facilitate likelihood computation in normalizing flows.

\subsection{Discrete Legendre Transform (DLT)}

DLT is defined as:
\begin{gather}
    X[k] = \sum_{n=0}^{N-1} \hat{P}_k(t_n)\, x[n], 
\end{gather}
where
\begin{gather}
    t_n = -1 + \frac{2n}{N-1}, \\[1ex]
    \quad P_0(t_n) = 1, \quad P_1(t_n) = t_n, \\[1ex]
    \quad P_{k+1}(t_n) = \frac{(2k+1)\, t_n\, P_k(t_n) - k\, P_{k-1}(t_n)}{k+1}, \quad k = 1,2,\ldots, N-2, \\[1ex]
     \quad \hat{P}_k(t_n) = \frac{P_k(t_n)}{\|P_k\|_2}, \quad \|P_k\|_2 = \sqrt{\sum_{n=0}^{N-1} \left[P_k(t_n)\right]^2}.
\end{gather}
In machine learning, Discrete Legendre Transform (DLT) has been applied to solving nonlinear integral equations, particularly Volterra–Fredholm–Hammerstein integral equations (V-F-H-IEs), using Legendre polynomials as activation functions \citep{LDNN2021}. Additionally, Legendre polynomial-based neural networks have been employed for heat and mass transfer analysis in non-Newtonian fluids within porous channels \citep{Khan2022} and for solving elliptic partial differential equations (PDEs) with enhanced accuracy and efficiency \citep{Yang2019}.

\subsection{Other Discrete Transforms in Machine Learning}

Beyond the above transforms, a range of other discrete transforms have been explored for machine learning applications:
\begin{itemize}
  \item \textbf{Discrete Hartley transform}: Offers the advantage of using only real arithmetic, eliminating the need for complex-valued computations required in DFT. It has been applied in computer vision for various image processing tasks \citep{https://doi.org/10.48550/arxiv.1810.04028,Wong2023, 9604984} and in NLP for handling long-context representations  \citep{giofré2023legalhnetmixinglegallongcontext}.
  \item \textbf{Fractional Fourier transform}: Extends the standard Fourier transform to partial-frequency domains, benefitting signal denoising and feature extraction \citep{Sahinuc2022, FrFT-denoising1, FrFT-denoising2}.
  \item \textbf{Walsh--Hadamard transform}: Offers binary orthogonal bases and fast computation, making it suitable for embedded systems and resource-constrained scenarios. WHT has been applied to vision tasks and integrated into neural networks \citep{WalshHadamard2022, Pan_2022_CVPR, Baldini2023}.
\end{itemize}


\section{Methods}\label{sec2}

\subsection{General Transform}

A major limitation of all the approaches above is their reliance on predefined, input-invariant operations, which may not optimally capture task-specific features and often require prior dataset knowledge, reducing their generality for multimodal or heterogeneous data. To address these issues, we propose General Transform (GT). 

Our definition of GT is as follows: 
\begin{gather}
    X[k] = \sum_{n=0}^{N-1} \left(\sum_{i=1}^m p_if_i[n, k] + \left(1-\sum_{i=1}^m p_i\right)f_{m+1}[n, k]\right)x[n]
\end{gather}
Here, $n$ and $k$ are indices, $x[n]$ represents the original sequence of length $N$, $X[k]$ is the transformed representation of $x[n]$, $m+1$ is the number of transforms to be included in the GT, and $f_i$ is the function for the $i$-th transform. The $p_i$ is a trainable parameter for the $i$-th transform, optimized alongside the network parameters during training. 

First, multiple transforms are selected to define the GT formula. During training, the weight of each transform ($p_i$) is optimized. Consequently, the GT mapping optimally converges to capture dataset features without requiring prior knowledge, dynamically adjusting the contribution of individual transforms, including their activation or deactivation, based on learned weights. The expressiveness of GT can be further enhanced by increasing the number of selected transforms. Moreover, the number of additional parameters introduced in this scheme corresponds directly to the number of included transformations. 

While the above formulation uses addition to blend the different discrete transforms, the GT framework can accommodate any parametric combination (e.g., sums, products, or more complex mappings) of multiple discrete transforms, as long as the resulting operator (1) smoothly interpolates between two or more transforms as a function of one or more blending parameters, (2) recovers standard transforms at specific parameter values.

\begin{figure*}[htbp]
    \centering
    \includegraphics[width=1\textwidth]{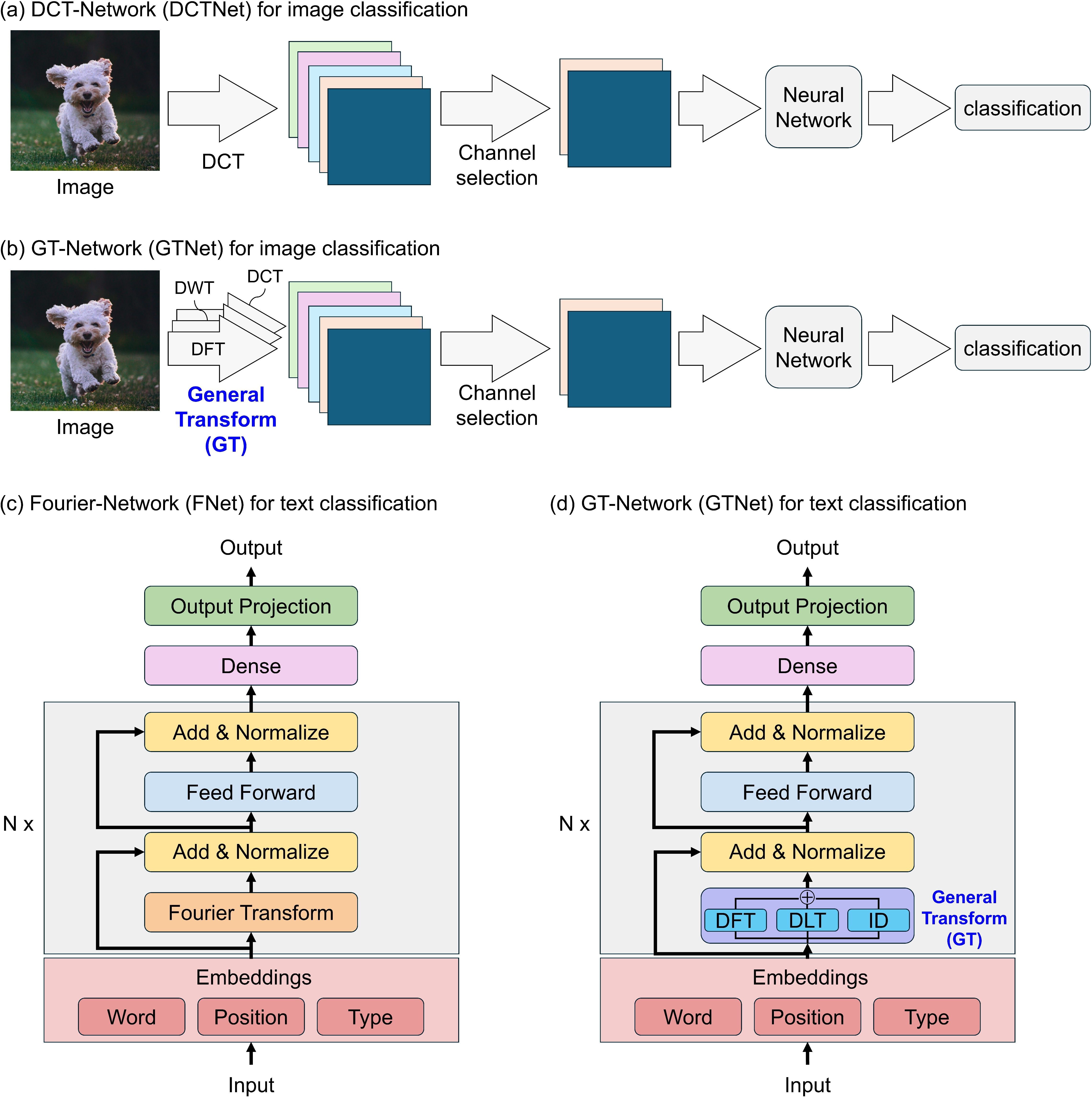}
    \caption{Image classification using RGB images as input, with DCT \citep{Xu2020} (a) and GT (b) for feature extraction. Text classification with DFT \citep{leethorp2022fnetmixingtokensfourier} (c) and GT (d) for token mixing.    }
    \label{fig1}
\end{figure*}

\subsection{Experiments}

\subsubsection{Computer Vision Task}

For the computer vision task, we implemented DFT (Eq. (1)), DCT (Eq. (2)), and DWT (Eq. (3)) as transformations, thereby defining GT for this task as:
\begin{gather}
    Y[k] = \sum_{n=0}^{N-1}\, \left(p_1\, \cos\Bigl(\tfrac{\pi k \,(n+0.5)}{N}\Bigr) + p_2\, e^{-\frac{j2\pi kn}{N}} + (1-p_1-p_2)\, H(k,n)\right)x[n].
\end{gather}
The final output is a parameterized summation of the real and imaginary components of $Y[k]$.
\begin{gather}
    X[k] = p_3\, Y[k]_{\mathrm{real}} + (1-p_3)\, Y[k]_{\mathrm{imaginary}}
\end{gather}
We replaced the DCT-based feature extraction used in \citep{Xu2020} with the GT (Fig. 1(a) and (b)). Following this work, we implemented three versions of each model, each employing a different number of frequency channels (24, 48, and 64) as input to the ResNet-50 \citep{He2016} architecture. In this experiment, the input image was first converted to the YCbCr color space, then divided into non-overlapping 8×8 blocks. A 2D DCT (DCTNet) or GT (GTNet) was applied to each block, resulting in 64 frequency coefficients. Depending on the model variant, only the first 24, 48, or all 64 coefficients were retained. These selected frequency channels were aggregated and normalized by channel-specific mean and standard deviation. The processed data was then fed into neural networks.

The models were trained from scratch using the ImageNet 2012 Large-Scale Visual Recognition Challenge (ILSVRC-2012) \citep{Deng2009} dataset. The cross-entropy loss function used as the objective function is given by:
\begin{equation}
    L = -\frac{1}{N} \sum_{i=1}^{N} \log\!\left(\frac{\exp(z_{i,y_i})}{\sum_{j=1}^{M}\exp(z_{i,j})}\right),
\end{equation}
where $N$ denotes the number of samples in the batch, $M$ represents the total number of classes, and $z_{i,j}$ is the unnormalized logit score corresponding to the $j$-th class of the $i$-th sample. The ground-truth class label for the $i$-th sample is denoted by $y_i \in \{1,\ldots,M\}$. In this context, the softmax probability assigned to the correct class $y_i$ is given as $q_{i,y_i} = \frac{\exp(z_{i,y_i})}{\sum_{j=1}^{M}\exp(z_{i,j})}$. During training, we adopted stochastic gradient descent (SGD) \citep{10.1007/978-3-7908-2604-3_16} with an initial learning rate of 0.1, a momentum of 0.9, and a weight decay of $10^{-4}$. Training proceeded for 80 epochs with a batch size of 150, and the learning rate was decayed by a factor of 0.1 every 31 epochs. The objective function $\mathcal{L}$ and the top-1 accuracy are used to evaluate model performance. The top-1 accuracy is defined as:
\begin{equation}
    A_{\text{Top-1}} = \frac{1}{N} \sum_{i=1}^{N} \mathds{1}(\hat{y}_i = y_i) \times 100\%,
\end{equation}
where $N$ denotes the number of samples in the batch, the indicator function $\mathds{1}(\cdot)$ returns 1 if its argument is true and 0 otherwise. Here, $y_i$ is the ground-truth label for the $i$-th sample, and $\hat{y}_i$ is the predicted label. The predicted label is computed as:
\begin{equation}
    \hat{y}_i = \mathop{\arg\max}_{j \in \{1,\ldots,M\}} \frac{\exp(z_{i,j})}{\sum_{k=1}^{M} \exp(z_{i,k})}.
\end{equation}
Because the softmax operation preserves the ordering of the logits—that is, the class with the highest logit will also have the highest softmax probability—we can simply choose the predicted label as:
\begin{equation}
    \hat{y}_i = \mathop{\arg\max}_{j \in \{1,\ldots,M\}} z_{i,j}.
\end{equation}
Thus, as the model outputs a probability distribution over the classes, the predicted label $\hat{y}_i$ is chosen as the class with the highest predicted probability.

For each model, the epoch with the highest validation $A_{\text{Top-1}}$ within the 80 training epochs is selected for comparison.

\subsubsection{Natural Language Processing Task}

For the NLP task, we implemented three transformation basis functions: the DFT (Eq. (1)), the DLT  (Eq. (5)), and an identity transform. Thus the GT used for the task is defined as:
\begin{gather}
      Y[k] \;=\;
  \sum_{n=0}^{N-1}
  \Bigl(
    p_{1}\,e^{-j\frac{2\pi kn}{N}}
    \;+\;
    p_{2}\,\hat{P}_{k}\!\left(t_{n}\right)
    \;+\;
    (1-p_{1}-p_{2})\,\delta_{k,n}
  \Bigr)
  \,x[n],
  \label{eq:gt-sum}
\end{gather}
where the $\delta_{k,n}$ is
\[
  \delta_{k,n} \;=\;
  \begin{cases}
    1, & \text{if } k = n,\\
    0, & \text{otherwise.}
  \end{cases}
\]
The final output is a parameterized summation of the real and imaginary components of $Y[k]$.
\begin{gather}
    X[k] = p_3Y[k]_{\mathrm{real}} + (1-p_3)Y[k]_{\mathrm{imaginary}}
\end{gather}
Then, we replaced the DFT-based token mixing proposed in \citep{leethorp2022fnetmixingtokensfourier} with GT (Fig. 1(c) and (d)). Here, each text sample is first tokenized using a SentencePiece model with a 32k-vocabulary \citep{kudo-richardson-2018-sentencepiece}. Then, the tokens are mapped to vectors through an embedding layer that includes token, positional, and segment embeddings. Within each encoder block, the self-attention sublayer of the Transformer is replaced by a two-dimensional DFT or GT applied across both the sequence and hidden dimensions. The processed data is then fed into feed-forward neural networks.

We evaluated the effectiveness of GT by fine-tuning pre-trained models. The model parameters were initialized using the pre-trained weights and subsequently fine-tuned on the SST-2 \citep{socher-etal-2013-recursive} and CoLA \citep{Warstadt2019} datasets, using either DFT (FNet) or GT (GTNet) for token mixing. GT’s parameters were initialized with a pure real-part DFT configuration (\(p_{1}=1,\; p_{2}=0,\; p_{3}=1\)), following the FNet setup.

During fine-tuning, the learning rate was set to increase to $10^{-5}$ by epoch 2 and gradually decayed to 0 by epoch 20. However, the primary analysis was focused on the first five epochs, as severe over-fitting was observed beyond this point. We used cross-entropy loss (Eq. (13)) as the training objective. The models were fine-tuned using the AdamW optimizer \citep{loshchilov2019decoupled} with a weight decay coefficient of 0.01. Specifically, we employed $\beta_1 = 0.9$, $\beta_2 = 0.999$ and $\text{eps} = 1\times10^{-6}$. A maximum sequence length of 512 tokens was used, with a batch size of 64. The fine-tuning process was repeated 10 times, and the $L$ and $A_{\text{Top-1}}$ were averaged for evaluation. For each model, the epoch that achieves the highest validation $A_{\text{Top-1}}$ within the 5 training epochs is chosen for comparison.

\begin{table}[h]
    \centering
    \caption{ImageNet classification results using DCTNet and GTNet with varying numbers of channel inputs to ResNet-50.}
    \setlength{\tabcolsep}{5pt} 
    \renewcommand{\arraystretch}{1.2} 
    \label{tab1}
    \begin{tabular}{|c|c|c|c|c|c|}
        \hline
        &  & \multicolumn{2}{c|}{\textbf{Training}} & \multicolumn{2}{c|}{\textbf{Validation}}  \\ \hline
        \textbf{Model} & \textbf{Channel} & \textbf{$L$} & $A_{\text{Top-1}}$ & \textbf{$L$} & $A_{\text{Top-1}}$  \\ \hline
        DCTNet & 24 & 0.9688 & 76.70 & 0.9388 & 76.53 \\ \hline
        GTNet & 24 & 0.9599 & 76.90 & 0.9377 & 76.62 \\ \hline
        DCTNet & 48 & 0.9637 & 76.87 & 0.9322 & 76.63 \\ \hline
        GTNet & 48 & 0.9358 & 77.43 & 0.9203 & 76.90 \\ \hline
        DCTNet & 64 & 0.9453 & 77.27 & 0.9246 & 76.67 \\ \hline
        GTNet & 64 & 0.9503 & 77.16 & 0.9268 & 76.72 \\ \hline
    \end{tabular}
\end{table}
\section{Results}\label{sec3}

\subsection{Computer Vision Task}

\begin{figure*}[htbp]
    \centering
    \includegraphics[width=1\textwidth]{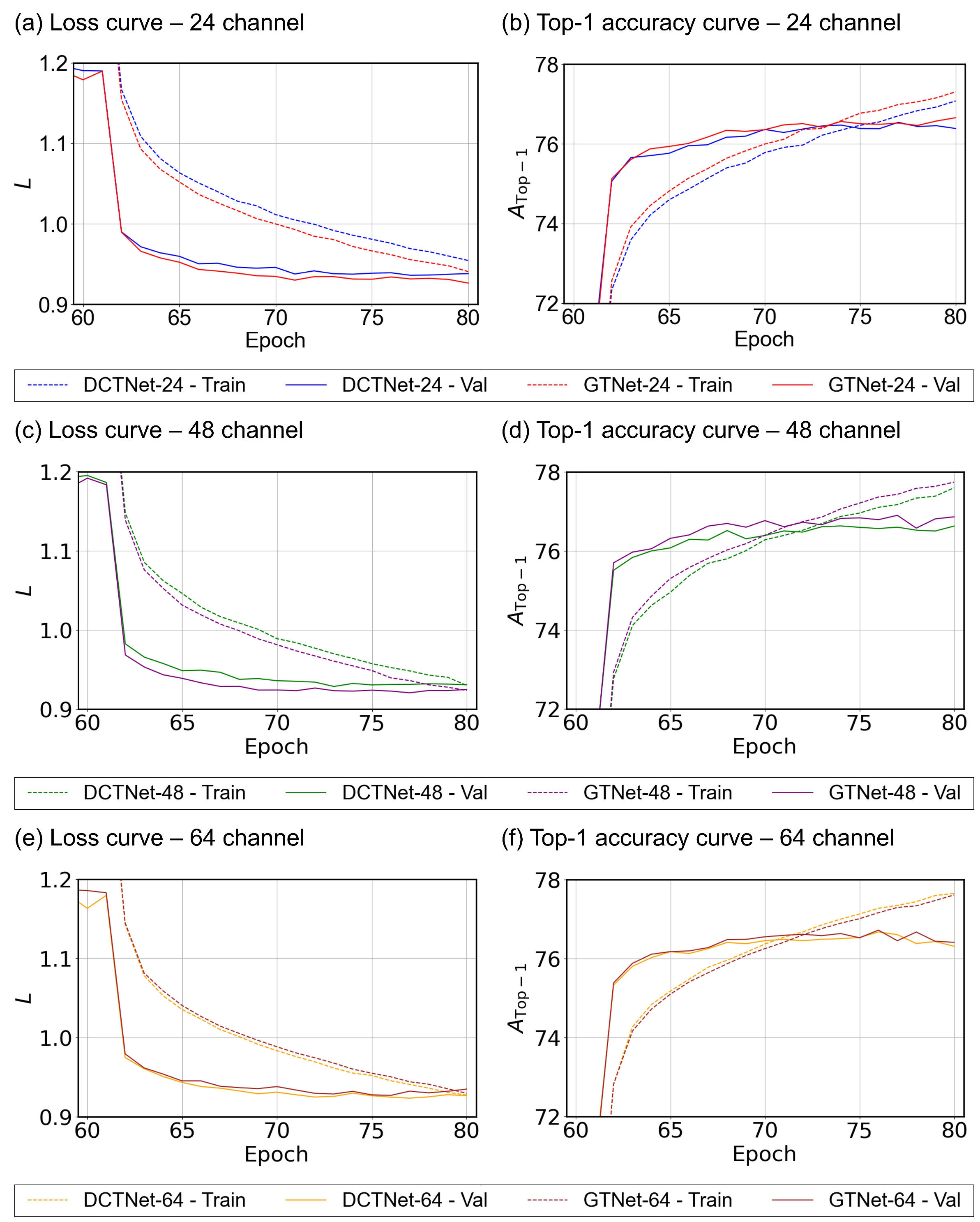}
    \caption{Loss and accuracy curves for image classification on the ImageNet 2012 dataset using DCTNet and GTNet, with 24 (a, b), 48 (c, d), and 64 (e, f) input channels.
    }
    \label{fig2}
\end{figure*}

For computer vision task, replacing DCT with the GT on the same network improved validation $A_{\text{Top-1}}$ and reduces $L$ on the ImageNet 2012 dataset classification.
Table 1 compares the performance of DCTNet and the proposed GTNet for image classification using ImageNet dataset. GTNet consistently outperforms DCTNet in validation $A_{\text{Top-1}}$, with an increase from  76.53\% to 76.62\% for the 24-channel configuration, 76.63\% to 76.90\% for the 48-channel configuration, and 76.67\% to 76.72\% for the 64-channel configuration. Additionally, the validation $L$ for GTNet with the 24-channel and 48-channel configurations is lower than their DCTNet counterparts: 0.9377 compared to 0.9388 for the 24-channel configuration, and 0.9203 compared to 0.9322 for the 48-channel configuration. Note that this improvement was achieved with the addition of only three parameters. Compared to the 25 million parameters of the ResNet-50 architecture, this additional computational cost is negligible.

Figure 2 presents the training and validation $L$ curves (left) and $A_{\text{Top-1}}$ curves (right) for DCTNet and GTNet across different frequency channel configurations: 24 (a, b), 48 (c, d), and 64 (e, f). In the $L$ curves (Fig. 2(a, c, e)), GTNet consistently exhibits lower validation loss than DCTNet, particularly at 24 channels  (Fig. 2(a), GTNet-24: red solid line; DCTNet-24: blue solid line) and  48 channels (Fig. 2(c), GTNet-48: purple solid line; DCTNet-48: green solid line), indicating improved generalization. However, as the number of channels increases to  64, the performance gap narrows (Fig. 2(e)), likely due to the larger network capacity diminishing the relative benefits of GTNet. In the $A_{\text{Top-1}}$ curves (Fig. 2(b, d, f)), GTNet consistently outperforms DCTNet across all channel configurations, maintaining a similar accuracy improvement at 24 (Fig. 2(b), GTNet-24: red solid line; DCTNet-24: blue solid line), 48 (Fig. 2(d), GTNet-48: purple solid line; DCTNet-48: green solid line), and 64 (Fig. 2(f), GTNet-64: brown solid line; DCTNet-48: yellow solid line) channels.

\begin{table}[tb]
    \centering
    \caption{Optimized parameters of General Transform.}
    \setlength{\tabcolsep}{5pt} 
    \renewcommand{\arraystretch}{1} 

    \newcommand{\pcol}[1]{\makebox[0.8cm][c]{#1}}

    \begin{tabular}{|c|c|c|c|c|c|c|c|c|c|}
        \hline
        & \multicolumn{3}{c|}{\textbf{Y-channel}} & \multicolumn{3}{c|}{\textbf{Cb-channel}} & \multicolumn{3}{c|}{\textbf{Cr-channel}} \\ \hline
        & \pcol{$p_1$} & \pcol{$p_2$} & \pcol{$p_3$} & \pcol{$p_1$} & \pcol{$p_2$} & \pcol{$p_3$} & \pcol{$p_1$} & \pcol{$p_2$} & \pcol{$p_3$} \\ \hline
        GTNet-24 & 0.84 & 0.15 & 0.65 & -1.85 & -0.21 & 0.18 & 1.94 & 0.18 & 0.16 \\ \hline
        GTNet-48 & -3.63 & -0.06 & 0.15 & 2.10 & 0.74 & 0.30 & 2.11 & 0.74 & 0.31 \\ \hline
        GTNet-64 & -0.90 & -1.68 & 2.20 & 2.93 & 1.63 & 0.66 & 3.12 & 1.71 & 0.61 \\ \hline
    \end{tabular}

    \label{tab2}
\end{table}

GT as defined in Eq. (11) recovers standard transforms at specific parameter values. However, our analysis of the optimized GT parameters reveals that the optimal mapping is not a standalone component transform and differs between the input channels and the model size (Table 2). For instance, the optimal values of $p_1$ and $p_2$ for the Y-channel of GTNet-24 are 0.84 and 0.15, respectively, indicating that a mixture dominated by DCT and DFT, with a minor contribution from DWT, yields optimal performance. In contrast, for the larger GTNet-48, the optimal $p_1$ and $p_2$ for the Y-channel are -3.63 and -0.06, respectively, suggesting that a combination favoring DCT and DWT, with a negligible influence from DFT, is preferable.
We also observed that, within the same model, the luminance (Y) channel has distinct $p_1$ and $p_2$ values compared to the chromatic (Cb, Cr) channels. This difference likely arises because the Y-channel captures more structural and high-frequency details compared to the chrominance channels (Cb and Cr), which primarily encode color information. Notably, GT effectively captures and adapts to these differences, learning distinct mappings for luminance and chrominance.

\begin{table}[b]
    \centering
    \caption{Fine-tuning results of FNet and GTNet on CoLA and SST-2 tasks.}
    \setlength{\tabcolsep}{4pt} 
    \renewcommand{\arraystretch}{1.2} 
    \begin{tabular}{|c|c|c|c|c|c|c|c|c|c|}
        \hline
        \textbf{} & \multicolumn{4}{c|}{\textbf{CoLA}} & \multicolumn{4}{c|}{\textbf{SST-2}} \\ \hline
        \textbf{} & \multicolumn{2}{c|}{\textbf{Training}} & \multicolumn{2}{c|}{\textbf{Validation}}  & \multicolumn{2}{c|}{\textbf{Training}} & \multicolumn{2}{c|}{\textbf{Validation}}  \\ \hline
         & $L$ & $A_{\text{Top-1}}$ & $L$ & $A_{\text{Top-1}}$ & $L$ & $A_{\text{Top-1}}$ & $L$ & $A_{\text{Top-1}}$ \\ \hline
        FNet-base& 0.4225 & 81.40 & 0.5875 & 72.48 & 0.1607 & 94.03 & 0.3876 & 87.58  \\ \hline
        GTNet-base& 0.4192 & 81.58 & 0.5624 & 74.04 & 0.1256 & 95.47 & 0.3835 & 88.50 \\ \hline
        FNet-large& 0.2828 & 89.49 &0.5336 & 77.75& 0.0825 & 97.29 & 0.3560 & 90.04  \\ \hline
        GTNet-large& 0.2821 & 89.54 & 0.5238 & 78.74 & 0.0834 & 97.26 &  0.3342 & 90.70  \\ \hline
    \end{tabular}
    \label{tab3}
\end{table} 
\subsection{Natural Language Processing Task}

GT also improved the model performance compared to DFT in natural language processing tasks. Table 3 summarizes the fine-tuning results of the pre-trained network, which employs DFT (FNet) and GT (GTNet) for token mixing, for both the base and large models with 83 million and 238 million parameters, respectively. 

\begin{figure*}[htbp]
    \centering
    \includegraphics[width=1\textwidth]{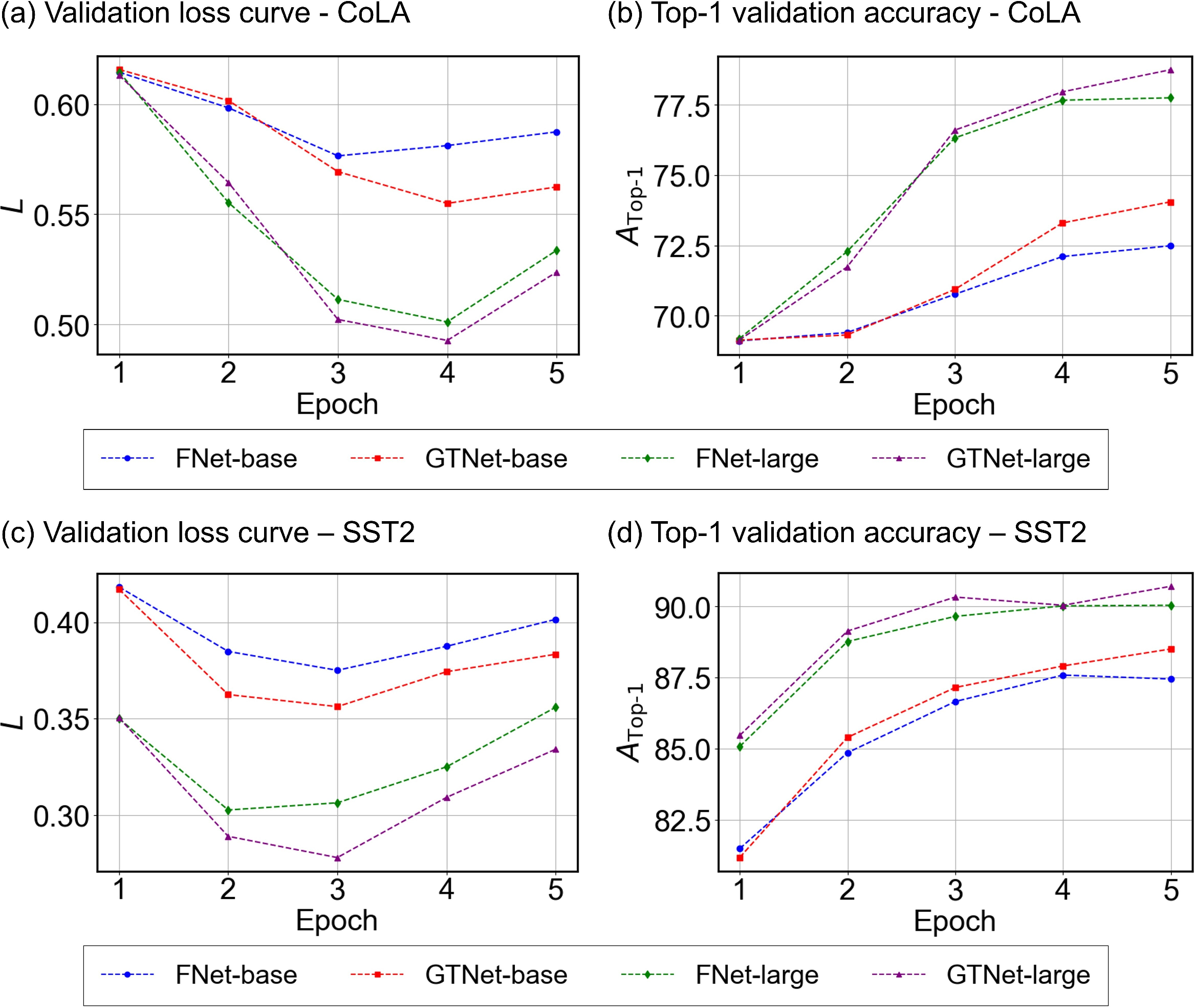}
    \caption{Validation loss ($L$) curve and top-1 accuracy ($A_{\text{Top-1}}$) curve for CoLA (a and b) and SST-2 (c and d) datasets.
    }
    \label{fig3}
\end{figure*}
For the CoLA task, GTNet consistently outperforms FNet in both validation $A_{\text{Top-1}}$ and $L$ across both model sizes. GTNet-base achieves a validation $A_{\text{Top-1}}$ of 74.04\%, compared to 72.48\% for FNet-base. For large models, GTNet-large achieves 78.74\%, surpassing the 77.75\% of FNet-large. GTNet also demonstrates lower validation $L$ compared to FNet across both base and large models. GTNet-base achieves a validation loss of 0.5624, compared to 0.5875 for FNet-base. For large models, GTNet-large achieves 0.5238, while FNet-large achieves 0.5336. 

For the SST-2 task, GTNet also demonstrates superior performance compared to FNet. The validation $A_{\text{Top-1}}$ for GTNet-base is 88.50\%, higher than the 87.58\% of FNet-base. For the larger models, GTNet-large achieves 90.70\%, outperforming the 90.04\% of FNet-large. GTNet consistently achieves lower validation $L$ as well, with GTNet-base achieving 0.3835, compared to 0.3876 for FNet-base. For the large models, GTNet-large achieves 0.3342, compared to 0.3560 for FNet-large. 

Figure 3 presents the average validation $L$ (a, c) and $A_{\text{Top-1}}$ (b, d) curves for FNet and GTNet on the CoLA and SST-2 datasets over the first five epochs of fine-tuning. 
In the CoLA validation $L$ curve (Fig. 3(a)), GTNet-large (purple) achieves a noticeably lower loss than FNet-large (green). GTNet-base (red) also outperforms FNet-base (blue). A similar trend is observed in the SST-2 validation $L$ curve (Fig. 3(c)), where GTNet-large (purple) achieves a lower $L$ than FNet-large (green). GTNet-base (red) consistently outperforms FNet-base (blue). For validation $A_{\text{Top-1}}$ on CoLA (Fig. 3(b)), GTNet-large (purple) outperforms FNet-large (green), while GTNet-base (red) consistently surpasses FNet-base (blue). On SST-2 validation $A_{\text{Top-1}}$ (Fig. 3(d)), GTNet-large (purple) maintains a performance lead over FNet-large (green). GTNet-base (red) consistently outperforms FNet-base (blue) throughout training, demonstrating superior accuracy across all epochs. 

\section{Extension to Quantum Computing Architectures}\label{sec4}

\subsection{Parameterized Quantum General Transform via Linear Combination of Unitaries}

The above General Transform (GT) can be further extended to quantum computing by leveraging the principle of Linear Combination of Unitaries (LCU) \citep{LCU1, LCU2, Kosugi_lcu}. This extension yields the Quantum General Transform (QGT), a quantum-native, trainable operator that performs adaptive feature mapping through a coherent superposition of multiple unitaries. Rather than selecting a single fixed unitary transformation \( U \), the QGT is defined as a variationally parameterized operator of the form:
\begin{equation}
U_{\mathrm{QGT}} := \sum_{i=0}^{m-1} p_i U_i
\end{equation}
Here, \( \{ U_i \}_{i=0}^{m-1} \) are known \( n \)-qubit unitary matrices, such as Quantum Fourier Transform (QFT) \citep{Shor, nielsen00}, Instantaneous Quantum Polynomial (IQP) \citep{Shepherd2009, Bremner2017achievingquantum},  or Clifford+T operators \citep{clifford+, nielsen00}, and \( \{ p_i \} \) are trainable, positive weights constrained by \( \sum_i p_i = 1 \). 

A typical LCU implementation introduces ancilla qubits prepared in a superposition state:
\begin{equation}
|\chi\rangle = \sum_i \sqrt{p_i} |i\rangle
\end{equation}
where \(|i\rangle\) is the computational basis state of the ancilla that acts as a selector for the corresponding unitary \(U_i\). 

Following the standard LCU framework, we define the multiplexed unitary:
\begin{equation}
\text{SELECT}(U) := \sum_i |i\rangle\langle i| \otimes U_i
\end{equation}

Post-selecting the ancilla back onto the state \( |\chi\rangle \) yields the effective operation:
\begin{equation}
\langle \chi | \text{SELECT}(U) | \chi \rangle = U_{\mathrm{QGT}},
\end{equation}
recovering the non-unitary operator via coherent projection. In physical implementations, amplitude amplification may be used to boost the success probability of this postselection. Thus, QGT acts as a learnable quantum module with learnable inductive bias over known quantum primitives.

\subsection{Quantum General Transform for Image Classification Tasks}

\subsubsection{Experiments}

We investigate the application of the Quantum General Transform (QGT) in the context of image classification, using the ImageNet2012 dataset as a benchmark \citep{Deng2009}. In this setting, QGT acts as a trainable, quantum-native transformation applied to image patches prior to classification. Following the experimental setup in \citep{Xu2020}, we substitute their DCT-based feature extraction pipeline with QGT of 3-qubit system, using 64 frequency channels as input to the ResNet-50 architecture. The classical input vectors are embedded into quantum Hilbert space through amplitude encoding, serving as the initial state for the QGT circuit. 

To explore the expressive capacity and trainability of the QGT under various quantum gate compositions, we test configurations comprising either two or four unitaries (Table 4). The unitary operators considered include the Quantum Fourier Transform (QFT) \citep{Shor, nielsen00}, Instantaneous Quantum Polynomial (IQP) circuits \citep{Shepherd2009, Bremner2017achievingquantum}, Clifford+T \citep{clifford+, nielsen00}, and parameterized Quantum Neural Networks (QNNs) \citep{StronglyEntLayers} (Fig. 4).

\begin{figure*}[htbp]
    \centering
    \includegraphics[width=1\textwidth]{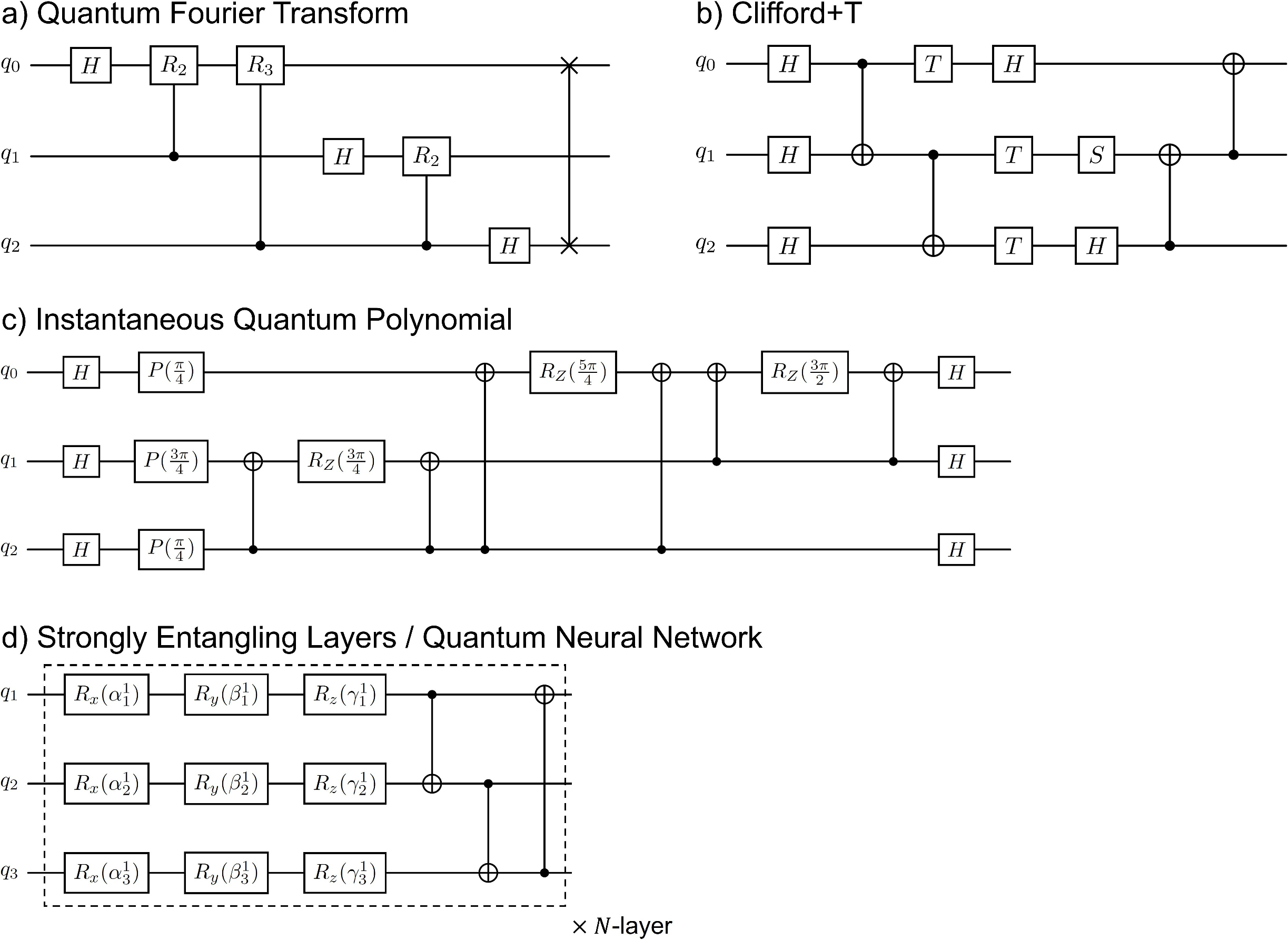}
    \caption{Different unitaries used for QGT: (a) Quantum Fourier Transform (QFT) where $R_k = \left(\begin{smallmatrix} 1 & 0 \\ 0 & e^{i 2\pi / 2^k} \end{smallmatrix}\right)$, (b) Clifford+T, (c) Instantaneous Quantum Polynomial (IQP) and (d) Quantum Neural Network (QNN).
    }
    \label{fig4}
\end{figure*}

\begin{table}[t]
\centering
\caption{Experimental Configurations for QGT-based ImageNet2012 Classification}
\label{tab4}
\begin{tabular}{|c|c|c|}
\hline
\makecell{\textbf{Experiment}\\\textbf{ID}} & \makecell{\textbf{\#LCU}\\\textbf{Unitaries}} & \textbf{Unitary Types Used} \\
\hline
S-1 & 2 & IQP, QFT \\
S-2 & 2 & IQP, Clifford+T  \\
S-3 & 2 & Clifford+T, QFT\\
S-4 & 4 & Clifford+T, QFT, IQP, QNN (5-layers)\\
\hline
\end{tabular}
\end{table}

\subsubsection{Results and Discussion}

\begin{figure*}[htbp]
    \centering
    \includegraphics[width=1\textwidth]{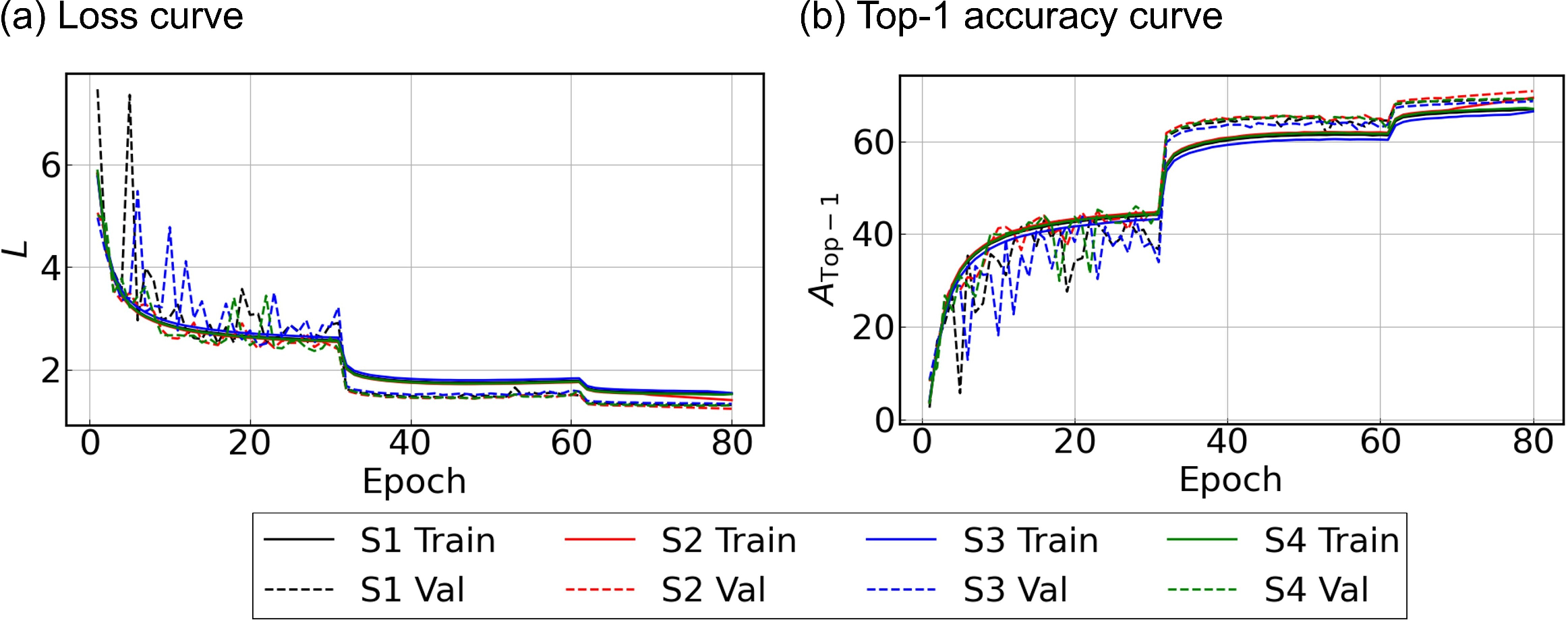}
    \caption{Training and validation loss ($L$) curves (a) and top-1 accuracy ($A_{\text{Top-1}}$) curves (b) for image classification on the ImageNet 2012 dataset using quantum general transform.
    }
    \label{fig5}
\end{figure*}

Figure 5 presents the training and validation performance for all four QGT-based configurations on the ImageNet2012 dataset. Figure 5(a) shows the evolution of the cross-entropy loss \(L\) over 80 training epochs, while Figure 5(b) illustrates the corresponding $A_{\text{Top-1}}$. Across all experimental settings (S1–S4), the QGT exhibits stable convergence, with monotonically decreasing training loss and improving accuracy, demonstrating consistent learnability.

Although the performance does not surpass that of classical General Transforms (GT), it is important to note that all hyperparameters—including optimizer settings and total training epochs—were inherited directly from the classical baseline, without any additional tuning for the quantum setting. Despite this, the QGT proves to be fully trainable on a large-scale, real-world classification task, underscoring its viability as a quantum-native preprocessing strategy.

Among the tested variants, S2 and S4—which incorporate Clifford+T unitaries—tend to yield slightly better generalization. Interestingly, S2 (comprising only two unitaries) outperforms S4 (which combines four unitaries including a parameterized QNN) in validation $A_{\text{Top-1}}$, suggesting that increasing the number of unitaries does not straightforwardly lead to better performance. This highlights the importance of careful circuit composition.

This study constitutes a preliminary exploration of the circuit design space for QGT in vision tasks. These results should be interpreted as a proof of concept, rather than an optimized system. Future work will focus on circuit architecture search, adaptive training protocols, and improved parameter initialization schemes to better exploit the quantum model capacity.

\section{Conclusion}\label{sec6}

In this study, we proposed General Transform (GT) for machine learning applications. GT generalizes the use of transforms in machine learning by allowing models to dynamically determine the optimal transform or combination of transforms based on the task and dataset. Through experiments on computer vision and natural language processing tasks, we demonstrated that GT consistently improves the performance of large-scale deep learning models. Notably, these gains are achieved with minimal overhead, requiring only a few additional parameters. The ability of GT to adapt to diverse datasets and tasks without requiring prior knowledge of their structure underscores its robustness and generalizability. These properties make GT a promising paradigm for more flexible and efficient transform-based learning in neural networks.

\section*{Acknowledgements}
This work was supported by JSPS KAKENHI under Grant-in-Aid for Early-Career Scientists No. JP24K16985 and JSPS KAKENHI under Grant-in-Aid for Transformative Research Areas No. JP22H05114. This study was partially carried out using the facilities of the Supercomputer Center, the Institute for Solid State Physics, the University of Tokyo and the TSUBAME4.0 supercomputer at the Institute of Science Tokyo. This work was partially supported by the Center of Innovations for Sustainable Quantum AI (JST Grant Number JPMJPF2221). The author acknowledges the contributions and discussions provided by the members of Quemix Inc.

\bibliographystyle{elsarticle-harv} 
\bibliography{manuscript}
\end{document}